\documentclass{ifacconf}

\usepackage{graphicx}      
\usepackage{natbib}        
\usepackage{amsmath}        
\begin{document}
\begin{frontmatter}

\title{Covariant fractional extension of the modified Laplace-operator used in 3D-shape recovery\thanksref{footnoteinfo}} 

\thanks[footnoteinfo]{}

\author[First]{Richard Herrmann} 

\address[First]{GigaHedron, Berliner Ring 80, D-63303 Dreieich (e-mail: herrmann@gigahedron.com)}

\begin{abstract}                
Extending  the Liouville-Caputo definition of a fractional derivative to  a nonlocal covariant generalization of arbitrary bound operators acting on multidimensional Riemannian spaces an appropriate approach for the 3D shape recovery of aperture afflicted 2D slide sequences is proposed. We demonstrate, that the step from a local to a nonlocal algorithm yields an order of magnitude in accuracy and by using the specific fractional approach an additional factor 2 in accuracy of the derived results.
\end{abstract}


\end{frontmatter}

\section{Introduction}
From a historical point of view, fractional calculus provides us with a set of axioms and methods to extend the concept of a derivative operator from integer order n to arbitrary order $\alpha$, where $\alpha$ is a real or complex value.  

\begin{equation}
\label{art14first}
{d^n \over dx^n} \rightarrow {d^\alpha \over dx^\alpha}
\end{equation}

In the sense of (\ref{art14first}) fractional calculus has been frequently applied in the area of image processing, see e.g. [\cite{fal94}], [\cite{ort03}], [\cite{spa09}].

Alternatively we may consider fractional calculus as a specific prescription to extend the definition of a local operator to the nonlocal case. In this lecture, we will present a covariant, multidimensional generalization of the fractional derivative definition, which may be applied to any bound operator on the Riemannian space.  

As a first application, we will propose  a  specific non-local extension of the modified local Laplace-operator, which is widely used in problems of image processing. We will especially compare the local to the nonlocal approach for 3D-shape recovery from a set of 2D aperture afflicted slide sequences, which may be obtained e.g. in confocal microscopy or autofocus algorithms [\cite{zer35}], [\cite{spe82}].

It will be shown, that a major improvement of results is achieved for the nonlocal version of the modified Laplace-operator.

\section{The generalized  fractional derivative}
We will propose a reinterpretation of the fractional calculus as a specific procedure for a non-local extension of arbitrary local operators.
For that purpose, we start with
the Liouville definition of the left and right  fractional integral [\cite{lio32}]:
\begin{equation}
{_\textrm{\tiny{L}}}I^\alpha \, f(x) \!=\! 
\begin{cases}
(I_{+}^\alpha f)(x) \!=\!  
\frac{1}{\Gamma(\alpha)}   
     \int_{-\infty}^x \!\!\!  d\xi \, (x-\xi)^{\alpha-1} f(\xi)
 \\
   (I_{-}^\alpha f)(x) \!=\! 
\frac{1}{\Gamma(\alpha)}  
     \int_x^\infty  \!\!\!  d\xi \, (\xi-x)^{\alpha-1} f(\xi)
 \\
\end{cases}
\end{equation} 
With a slight modification of the fractional parameter $\alpha=1-a$, where $\alpha$ is in the interval  $0 \leq \alpha \leq 1$. Consequently for the limiting case $\alpha=0$ $I_{+}$ and $I_{-}$ both coincide with the unit-operator and for $\alpha=1$  $I_{+}$ and $I_{-}$ both correspond to  the standard integral operator.

$I_{+}$ and $I_{-}$  may be combined to define a regularized Liouville integral [\cite{her11}]:
\begin{eqnarray}
I^\alpha \, f(x) \ &=& 
({1 \over 2} (I_{+}^\alpha + I_{-}^\alpha) f)(x) \\  
&=& \frac{1}{\Gamma(\alpha)}   
     \int_{0}^{\infty} \!\!\!  d\xi \, \xi^{\alpha-1} {f(x+\xi)+f(x-\xi) \over 2}\\
&=& \frac{1}{\Gamma(\alpha)}   
     \int_{0}^{\infty} \!\!\!  d\xi \, \xi^{\alpha-1} \hat{s}(\xi) f(x)
\end{eqnarray} 
where we have introduced the symmetric shift-operator:
\begin{equation}
\hat{s}(\xi)f(x) = {f(x+\xi)+f(x-\xi) \over 2} 
\end{equation} 
The regularized fractional Liouville-Caputo derivative may now be defined as:
\begin{eqnarray}
\partial_x^\alpha \, f(x)  &=& I^\alpha  \partial_x f(x) \\
 &=& \left(
\frac{1}{\Gamma(\alpha)}   
     \int_{0}^{\infty} \!\!\!  d\xi \, \xi^{\alpha-1} \hat{s}(\xi)\right)  \partial_x f(x)\\
 &=&
\frac{1}{\Gamma(\alpha)}   
     \int_{0}^{\infty} \!\!\!  d\xi \, \xi^{\alpha-1} {f^{'}(x+\xi)+f^{'}(x-\xi) \over 2}\\
 &=&
\frac{1-\alpha}{\Gamma(\alpha)}   
     \int_{0}^{\infty} \!\!\!  d\xi \, \xi^{\alpha-1} {f(x+\xi)-f(x-\xi) \over 2 \xi}
 \end{eqnarray} 
with the abbreviation $\partial_x f(x) = f^{'}(x)$. This definition of a fractional derivative coincides with Feller's [\cite{fel52}] definition ${_\textrm{\tiny{F}}}\partial_x(\theta)$ for the special case $\theta=1$. 

We may interpret $I^\alpha$ as a non-localization operator, which is applied to the local derivative operator to determine a specific non-local extension of the same operator. Therefore the fractional extension of the derivative operator is separated into a sequential application of the standard derivative followed by a non-localization operation. The classical interpretation  of a fractional integral is changed from the inverse operation of a fractional derivative to a more general interpretation of a non-localization procedure, which may be easily interpreted in the area of image processing as a blur effect. 

This is a conceptual new approach, since it may be easily extended to other operators e.g. higher order derivatives or space dependent operators, e.g.  
for $\partial^2_x$ we obtain:
\begin{eqnarray}
(\partial_x^2) ^\alpha  &f(x)&  = I^\alpha  \partial_x^2 f(x) \\
 &=& \left(
\frac{1}{\Gamma(\alpha)}   
     \int_{0}^{\infty} \!\!\!  d\xi \, \xi^{\alpha-1} \hat{s}(\xi)\right)  \partial_x^2 f(x)\\
 &=& 
\frac{1}{\Gamma(\alpha)}   
     \int_{0}^{\infty} \!\!\!  d\xi \, \xi^{\alpha-1} {f^{''}(x+\xi)+f^{''}(x-\xi) \over 2}\\
&=&
\frac{1-\alpha}{\Gamma(\alpha)}   
     \int_{0}^{\infty} \!\!\!  d\xi \, \xi^{\alpha-1} {f^{'}(x+\xi)-f^{'}(x-\xi) \over 2 \xi}\\
 &=& 
\frac{2-\alpha}{2 \Gamma(\alpha)}   
     \int_{0}^{\infty} \!\!\!  d\xi \, \xi^{\alpha-1} {f(x+\xi)- 2 f(x) + f(x-\xi) \over \xi^2} \nonumber \\
\end{eqnarray} 
which is nothing else but the Riesz [\cite{rie49}] definition of a fractional derivative.

Therefore we define the following  fractional extension of a local operator ${_\textrm{\tiny{local}}}\hat{O}$ to the non-local case
\begin{equation}
\label{gen}
{_\textrm{\tiny{nonlocal}}}\hat{O}^\alpha  \, f(x) = I^\alpha  {_\textrm{\tiny{local}}}\hat{O} f(x) 
\end{equation} 

as the covariant generalization of the Liouville-Caputo fractional derivative to arbitrary operators on $\mathcal{R}$.

This definition may be easily extended to the multidimensional case, interpreting the variable $\xi$ as a measure of distance.

In two dimensions, with
\begin{equation}
\xi = \sqrt{\xi_1^2+\xi_2^2}
\end{equation} 
and with
\begin{eqnarray}
&\hat{s}(\xi_1,\xi_2)f(x,y) = \hat{s}(\xi_1)\hat{s}(\xi_2)f(x,y) \\
&=  {1 \over 4} \left(
f(x+\xi_1,y+\xi_2)+f(x-\xi_1,y+\xi_2)\nonumber \right. \\
&  \,\,\,\left. +f(x+\xi_1,y-\xi_2)+f(x-\xi_1,y-\xi_2) \right)
\end{eqnarray} 
$I^\alpha(x,y)$ explicitly reads:
\begin{eqnarray}
I^\alpha (x,y)
&=& 
\frac{1}{2^{a-2}\Gamma(\alpha/2)^{2} \sin(a \pi/2)} \times \nonumber \\   
   &&  \int_{0}^{\infty}\!\!\!\!\! d\xi_1  \int_{0}^{\infty} \!\!\!\!\!  d\xi_2 \, (\xi_1^2+\xi_2^2)^{\frac{1}{2}(\alpha-2)} \hat{s}(\xi_1,\xi_2) \nonumber \\
& & \qquad \qquad \qquad 0 \leq \alpha \leq 2 
  \end{eqnarray} 
which is normalized such, that the eigenvalue spectrum for: 
\begin{equation}
I^\alpha f(x,y) = \kappa f(x,y)
  \end{equation} 

with the eigenfunctions $f(x,y) = \exp^{i k_1 x + i k_2 y}$
follows as:

\begin{equation}
\kappa  = (k_1^2 + k_2^2)^{-\alpha/2}
 \end{equation}

It should be noted, that the validity range for $\alpha$ spans from $0 \leq \alpha \leq 2$, since we deal with a two-dimensional problem. Obviously within the framework of signal processing, the non-localization operator may  be interpreted as a low-pass filter.

In the following sections, we will use this operator for a well defined  extension of  the standard algorithm used for 3D-shape recovery from aperture afflicted 2D-slide sequences to a generalized, fractional nonlocal version, which results in a very stable procedure with drastically reduced errors.  

We will first present the minimal standard method and its
limitations in the next section.

\begin{figure}
\begin{center}
\includegraphics[width=8.4cm]{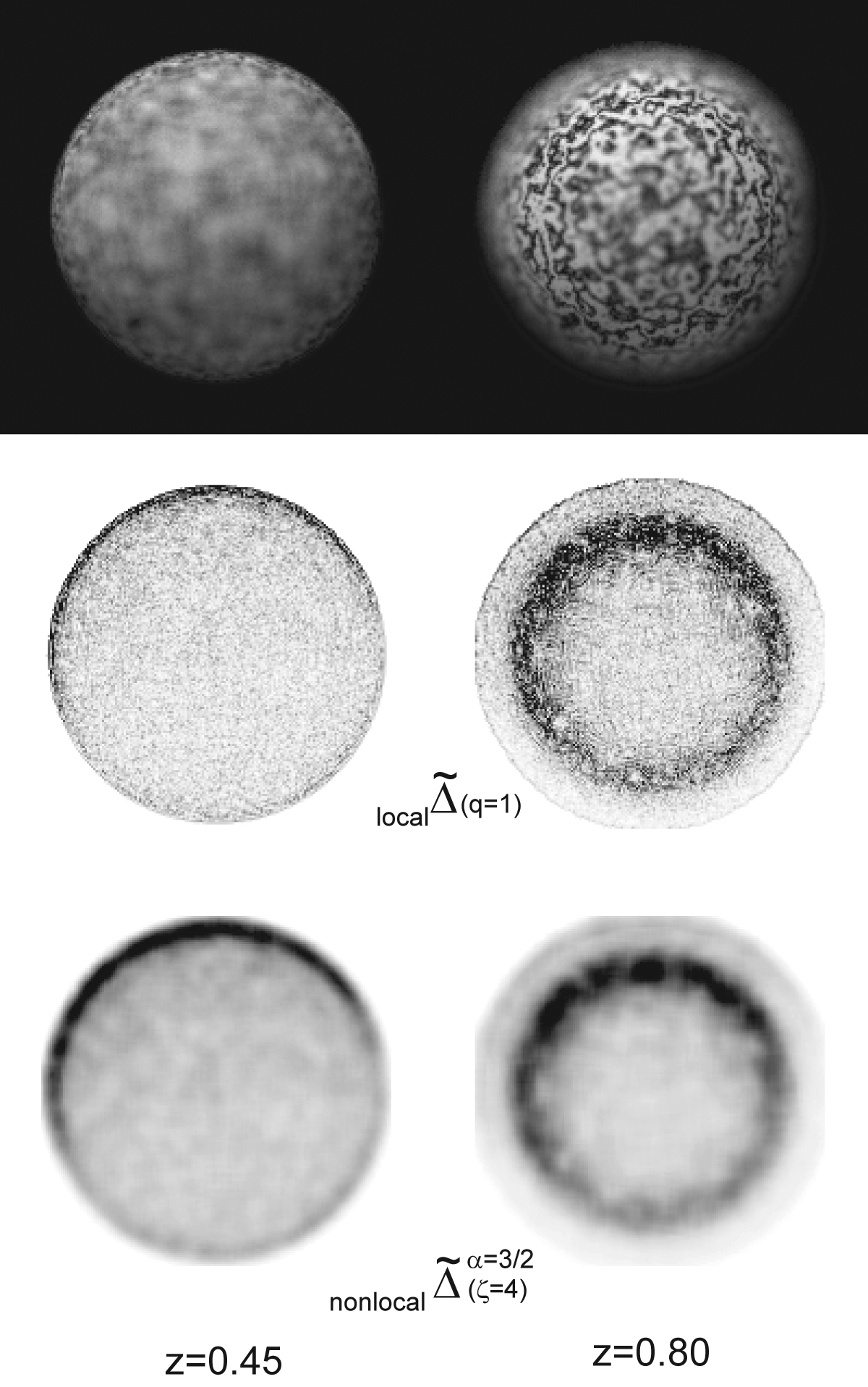}    
\caption{Application of the local and nonlocal modified Laplace-operator to 2 different original slides (left column $z=0.45$ and right column $z=0.8$) from a 2D-slide sequence $z \in \{0,1\}$. From top to bottom original slide, result of local modified Laplacian from (\ref{lap2})  ${_\textrm{\tiny{local}}}\tilde{\Delta}(q=1)$   and result of nonlocal operator are shown. }
\label{fig1}
\end{center}
\end{figure}

\section{The local approach}

In a  set $\{ p_i(z_i), i=1,...,N\}$  of $N$ 2D-slides with increasing focal distance $z_i$,
$
\{ z_i , \forall z_i: z{_\textrm{{min}}} \leq z_i \leq z{_\textrm{{min}}},  i=1,...,N\}
$
every slide contains areas with focused as well as defocused parts of the specimen considered. In the first row of Fig. \ref{fig1} we present two examples from a slide-sequence of a spherical object with radius $r=1$ located at $z=0$ in the x,y-plane, where the focal plane was chosen to be $z=0.45$ and $z=0.8$ respectively.

For a 3D-shape recovery in a first step for a given slide the parts being in focus have to be extracted. For a textured object, areas in focus are dominated by a larger amount of high frequency contributions, while for out of focus parts  mainly the low frequency    
amount of texture survives.

Consequently an appropriate operator to determine the high-frequency domains is the modified Laplacian ${_\textrm{\tiny{local}}}\tilde{\Delta}$ given e.g. by: 
\begin{equation}
({_\textrm{\tiny{local}}}\tilde{\Delta} \,\,  f)(x,y) \!=\!
(|{\partial^2 \over \partial_x^2}| + |{\partial^2 \over \partial_y^2}|
) 
f(x,y)
\end{equation} 
where $||$ denotes the absolute value. 

In the discrete case with a symmetrically discretized function $f_{ij}$  on a rectangular domain $x{_\textrm{min}} \leq x \leq x{_\textrm{max}}$ and $y{_\textrm{min}} \leq y \leq y{_\textrm{max}}$:
\begin{eqnarray}
f_{ij} &=& f(x{_\textrm{min}} +  i h, y{_\textrm{min}} +  j h) \\ 
&& \quad i=0,...,i_{_\textrm{max}},\, j=0,...,j{_\textrm{max}} \nonumber
\end{eqnarray}
with stepsize $h$ in both x- and y-direction, the same operator is given by:
\begin{eqnarray}
\label{lap2}
({_\textrm{\tiny{local}}}\tilde{\Delta}(q) &&f)_{ij} =  \nonumber \\
&&
|{ f_{i+q,j} \!- 2  f_{i,j} +  f_{i-q,j} \over (q h)^2}| \!+\! |{ f_{i,j+q} \!- 2  f_{i,j} +  f_{i,j-q} \over (q h)^2}|\nonumber \\
&& \quad\quad i=q,...,i_{_\textrm{max}}-q,\, j=q,...,j{_\textrm{max}}-q
\end{eqnarray} 
and $0$ elsewhere, where the free parameter $q$ has to be chosen according to the Nyquist-Shannon sampling theorem [\cite{sha49}] to be of order of the inverse average wavelength $\omega$ of the texture applied to the object considered
\begin{equation}
q h \approx 2/\omega
\end{equation} 
a requirement, which can be fulfilled only locally for random generated textures and for regular textures on curved surfaces respectively.

An application of the modified Laplacian to every slide in a set $\{p_k(z_k)\}$ leads to a set of  intensity values $\{\rho_{ij}(q,z_k)\}$ at a given pixel-position at $ij$:
\begin{equation}
\label{dloc}
{_\textrm{\tiny{local}}}\tilde{\Delta}(q) \{p_{ij,k}(z_k)\} = \{\rho_{ij}(q,z_k)\} 
\end{equation}
In the second row of Fig. \ref{fig1} the result of an application of the discrete modified Laplacian with $q=1$ to the original slides presented in the first row, is demonstrated.

It is assumed, that for a fixed $q$   a maximum  exists in $\rho_{ij}(q,z_k)$ for a given $\tilde{k}$. A parabolic fit of $\{\rho_{ij}(q,z_k)\}$ near $\tilde{k}$ helps to determine the position ${_\textrm{\tiny{opt}}}z(q)$, where $\rho_{ij}(q,z)$ is maximal:
\begin{equation}
\label{locz}
{_\textrm{\tiny{opt}}}z(q) = 
\tilde{k} -
\frac{1}{2}
{ 
\rho_{ij}(q,z_{{\tilde{k}+1}})
-\rho_{ij}(q,z_{{\tilde{k}-1}})
\over 
\rho_{ij}(q,z_{{\tilde{k}+1}})
-2 \rho_{ij}(q,z_{{\tilde{k}}})
+\rho_{ij}(q,z_{{\tilde{k}-1}})
}
\end{equation}

\begin{figure}
\begin{center}
\includegraphics[width=8.4cm]{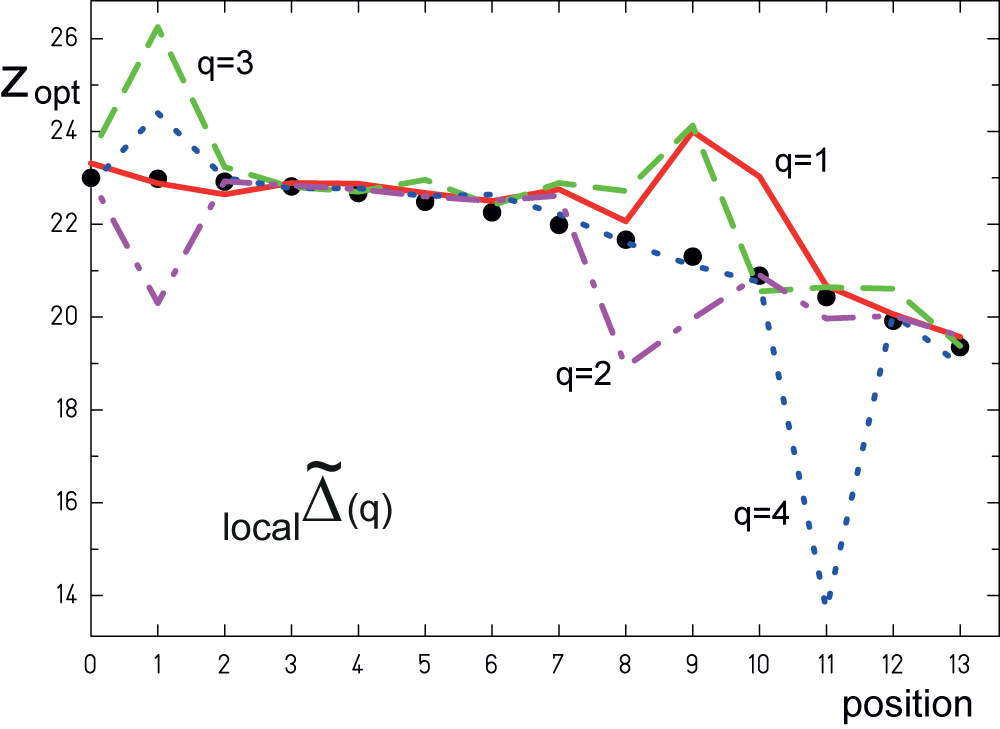}    
\caption{Comparison of recovered positions with original height positions for a set of points along the y-axis. Black circles mark the correct positions, lines represent recovered positions based on the local modified  Laplacian ${_\textrm{\tiny{local}}}\tilde{\Delta}(q)$ from 
(\ref{lap2})
for different values of step-size $q$. It should be noted, that there is no  fixed value of $q$, which uniquely may be used to determine all positions. There are drop outs for every curve. Errors are listed in Table \ref{art14tab1}. 
  }
\label{fig2}
\end{center}
\end{figure}

In the center row of Fig. \ref{fig1} we present the result of the application of (\ref{lap2}) onto the original slides. The gray-level indicates the intensity values $\{\rho_{ij}(q,z_k)\}$ for $z_k = 0.45$ and $z_k = 0.8$ respectively.   In Fig \ref{fig2} recovered $z{_\textrm{\tiny{opt}}}(q)$ along the positive y-axis are compared for different q-values with the original height-values. 

Obviously there is no unique optimum choice for q, which works for all positions simultaneously. The proposed simple local approach is not very effective, instead it generates drop outs as a result of an interference of varying texture scaling with the fixed step size q.       
For a realistic treatment of 3D-shape recovery a more sophisticated procedure is necessary.

Consequently a nonlocal approach, which weights the different contributions for a varying step size is a promising and well defined approach. Indeed it will enhance the quality of the results significantly, as will be demonstrated in the next section.

\section{The nonlocal approach}

The generalized fractional approach extends the above presented local algorithm. The nonlocal modified Laplacian according to  (\ref{gen}) is given by :
\begin{equation}
\label{genL2}
\begin{split}
{_\textrm{\tiny{nonlocal}}}&\tilde{\Delta}^\alpha(q)  \, f(x,y) =
( I^\alpha  {_\textrm{\tiny{local}}}\tilde{\Delta}(q)) f(x,y) \\
 &=  
\frac{1}{2^{a-2}\Gamma(\alpha/2)^{2} \sin(a \pi/2)}\, \times  \\   
   &  
 \int_{0}^{\infty}\!\!\!\!\!\!\! d\xi_1  \int_{0}^{\infty} \!\!\!\!\!\!\!  d\xi_2 \, (\xi_1^2+\xi_2^2)^{\frac{1}{2}(\alpha-2)} \hat{s}(\xi_1,\xi_2)
{_\textrm{\tiny{local}}}\tilde{\Delta}(q)    f(x,y) 
\end{split}
\end{equation} 

Therefore we obtain a well defined two step procedure. First, the local operator is applied, followed by the nonlocalization integral. 

In the discrete case, applying the nonlocal Laplacian to every slide in a slide set, the first step is therefore identical with (\ref{dloc}) and yields a set of intensity values
 $\{\rho_{ij}(q,z_k)\}$ at a given pixel-position at $ij$.
An application of the discrete version of $I^\alpha$ then leads to: 

\begin{equation}
\label{genL3}
\begin{split}
{_\textrm{\tiny{nonlocal}}}&\tilde{\Delta}^\alpha(q)  \, p_{ij,k}(z_k) = 
 I^\alpha  {_\textrm{\tiny{local}}}\tilde{\Delta}(q) \, p_{ij,k}(z_k) \\
 &=  
 I^\alpha  \rho_{ij}(q,z_k) \\
& = 
\frac{1}{2^{a-2}\Gamma(\alpha/2)^{2} \sin(a \pi/2)}\, \times  \\   
   &  
 \int_{0}^{\zeta}\!\!\! d\xi_1  \int_{0}^{\zeta} \!\!\!  d\xi_2 \, (\xi_1^2+\xi_2^2)^{\frac{1}{2}(\alpha-2)} \hat{s}(\xi_1,\xi_2)
\rho_{ij}(q,z_k)\\
&=\tilde{\rho}^\alpha_{ij}(q,z_k)
\end{split}
\end{equation} 

where we have introduced a cutoff $\zeta$, which limits the integral on the finite domain of pixel values. If we interpret the intensity values as constant function values at position $ij$ with size $h$, the integration may be performed fully analytically. In the appendix we have listed the resulting matrix-operator for $\zeta=4$. 

The resulting nonlocal intensities $\tilde{\rho}^\alpha_{ij}(q,z_k)$ are presented in
the lower row of Fig. \ref{fig1}. The nonlocal approach reduces the granularity of the local operator and a more smooth behaviour of intensities results.

Since this is the only modification of the local approach, the recovery of the height information for every pixel is similar to (\ref{locz})  

\begin{equation}
{_\textrm{\tiny{opt}}}z(\alpha, q) = 
\tilde{k} -
\frac{1}{2}
{ 
\tilde{\rho}^\alpha_{ij}(q,z_{{\tilde{k}+1}})
-\tilde{\rho}^\alpha_{ij}(q,z_{{\tilde{k}-1}})
\over 
\tilde{\rho}^\alpha_{ij}(q,z_{{\tilde{k}+1}})
-2 \tilde{\rho}^\alpha_{ij}(q,z_{{\tilde{k}}})
+\tilde{\rho}^\alpha_{ij}(q,z_{{\tilde{k}-1}})
}
\end{equation}

\begin{figure}
\begin{center}
\includegraphics[width=8.4cm]{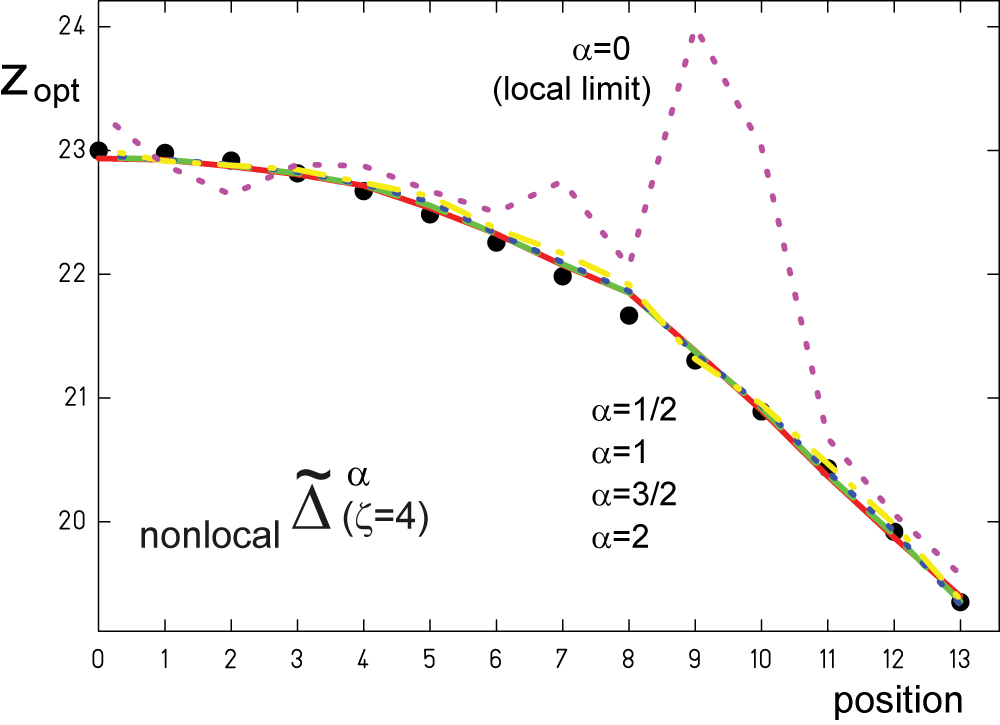}    
\caption{Comparison of recovered positions with original height positions for a set of points along the y-axis. Black circles mark the correct positions, lines represent recovered positions based on the nonlocal modified  Laplacian ${_\textrm{\tiny{local}}}\tilde{\Delta}(q)$ from 
(\ref{genL3})
for different values of the fractional parameter $\alpha$. The algorithm is very stable against a variation of $\alpha$. In the limit  $\alpha=0$ the nonlocal approach reduces to the local scenario. 
Errors are listed in Table \ref{art14tab1}.
  }
\label{fig3}
\end{center}
\end{figure}

In Fig. \ref{fig3} results are plotted for different values of $\alpha$.

\begin{table}
\caption{
\label{art14tab1}
Comparison of rms-errors in $\%$ for nonlocal modified Laplacian from (\ref{genL3}) for different $\alpha$  to local approach from (\ref{dloc}) with varying q in the last column.  
}
\begin{tabular}{l|lllll|l}
\hline\noalign{\smallskip}
$\zeta$& $\alpha=2.0$ & $\alpha=1.5$ & $\alpha=1.0$ & $\alpha=0.5$ &$\alpha=0.0$ & q \\ 
\noalign{\smallskip}\hline\noalign{\smallskip}
1& 0.277 & 0.265 &0.263 & 0.405 & 1.929 & 1.929 \\ 
2& 0.227 & 0.225 &0.225 & 0.301 & 1.929 & 2.496 \\ 
3& 0.211 & 0.211 &0.216 & 0.272 & 1.929 & 3.702 \\ 
4& 0.182 & 0.187 &0.198 & 0.251 & 1.929 & 3.702 \\ 
5& 0.145 & 0.151 &0.172 & 0.235 & 1.929 & 3.945 \\ 
6& 0.134 & 0.136 &0.158 & 0.223 & 1.929 & 6.131 \\ 
7& 0.137 & 0.138 &0.155 & 0.215 & 1.929 & 2.832 \\ 
8& 0.146 & 0.143 &0.154 & 0.205 & 1.929 & 4.228  \\
\hline\noalign{\smallskip}
\end{tabular}
\end{table} 

In Table \ref{art14tab1} a listing of errors is given for the local and the nonlocal algorithm presented. 

We may conclude, that the nonlocal approach is very robust and stable in a wide range of $\alpha$ and $\zeta$ values respectively. 
We gain one order of magnitude in accuracy using the nonlocal modified Laplacian. An additional  factor 2 in accuracy is obtained if we chose the optimal fractional $\{\alpha, \zeta\}$ parameter set.  

\hspace{1 cm}

\section{Appendix}
The discrete version of the nonlocalization operator $I^\alpha$ from (\ref{genL3}) may be interpreted as a matrix operation $M(\alpha)$ on $f_{ij}$:
\begin{equation} 
I^\alpha f_{ij} = M(\alpha) f_{ij}
\end{equation} 
$M(\alpha)$ is  a quadratic $(2 \zeta +1) \times (2 \zeta +1)$ matrix with the symmetry properties
\begin{equation} 
\begin{split}
M(\alpha)_{i,j} = M(\alpha)_{-i,j} = M(\alpha)_{-i,-j} &= M(\alpha)_{i,-j}=M(\alpha)_{j,i}\\
& i,j \leq \zeta  
\end{split}
\end{equation} 
Setting the normalization condition $M(\alpha)_{00}=1$  the integral may be solved analytically for stepwise constant pixel values $p_{ij}$.  As an example, we present the 
fourth quadrant of $M(\alpha)$ for $\zeta=4$ in units $h$:  
{\small{{
\begin{eqnarray}
M(2) &=&
\begin{pmatrix} 
  1.& 1.& 1.& 1.& 1.\\
  1.& 1.& 1.& 1.& 1.\\
  1.& 1.& 1.& 1.& 1.\\
  1.& 1.& 1.& 1.& 1.\\
  1.& 1.& 1.& 1.& 1.
\end{pmatrix} 
 \\ 
M(3/2) &=&
\begin{pmatrix} 
  1.             & 0.570351& 0.400990& 0.326971& 0.283027\\
  0.570351  & 0.478687& 0.379117& 0.318443& 0.278761\\
  0.400990  & 0.379117& 0.336822& 0.298160& 0.267640\\
  0.326971  & 0.318443& 0.298160& 0.274801& 0.253092\\
  0.283027  & 0.278761& 0.267640& 0.253092& 0.237922
  \end{pmatrix} 
\nonumber \\
M(1) &=&
\begin{pmatrix} 
  1.            & 0.294441& 0.143268& 0.094982& 0.071095\\
  0.294441 & 0.205559& 0.127951& 0.090073& 0.068963\\
  0.143268 & 0.127951& 0.100830& 0.078927& 0.063559\\
  0.094982 & 0.090073& 0.078927& 0.067014& 0.056825\\
  0.071095 & 0.068963& 0.063559& 0.056825& 0.050208
 \end{pmatrix} 
\nonumber
\end{eqnarray}
\nonumber
}}}
{\small{{
\begin{eqnarray}
M(1/2) &=&
\begin{pmatrix} 
  1.           & 0.116147& 0.038486& 0.020685& 0.013374\\
  0.116147& 0.066866& 0.032440& 0.019096& 0.012776\\
  0.038486& 0.032440& 0.022637& 0.015652& 0.011301\\
  0.020685& 0.019096& 0.015652& 0.012237& 0.009550\\
  0.013374& 0.012776& 0.011301& 0.009550& 0.007929
 \end{pmatrix} 
\nonumber\\
M(0) &=&
\begin{pmatrix} 
  1.           & 0.& 0.& 0.& 0.\\
  0.           & 0.& 0.& 0.& 0.\\
  0.           & 0.& 0.& 0.& 0.\\
  0.           & 0.& 0.& 0.& 0.\\
  0.           & 0.& 0.& 0.& 0.
  \end{pmatrix} 
\end{eqnarray}
}}}
Obviously there is a smooth transition from a local ($\alpha=0$) to a more and more nonlocal operation, which in the limiting case  ($\alpha=2$) may be interpreted as the result of the use of a pinhole camera with finite hole radius $\zeta$. 

\begin{ack}
We thank A. Friedrich and G. Plunien from TU Dresden, Germany for useful discussions. 
The original 2D-slide sequence, two examples shown in the topmost  row in Fig {\ref{fig1}}, was generated using povray [\cite{pov11}].
\end{ack}


\begin{thebibliography}{99}  
\bibitem[{Falzon(1994)}]{fal94} 
Falzon, F.  and Giraudon, G.  (1994). 
\emph{Singularity analysis and derivative scale-space} in 
Proceedings
CVPR '94, IEEE Computer Society Conference,245--250..
\bibitem[{Feller(1952)}]{fel52} 
Feller, W.  (1952). 
\emph{On a generalization of Marcel Riesz' potentials and the semi-groups generated by them} 
Comm. Sem. Mathem. Universite de Lund, 72--81.
\bibitem[{Herrmann(2011)}]{her11}  
Herrmann, R.  (2011). 
\emph{Fractional Calculus - An introduction for physicists}
World Scientific Publishing, Singapore.
\bibitem[{Liouville(1832)}]{lio32}
Liouville, J. (1832).
\emph{Sur le calcul des differentielles $\acute{\text{a}}$ indices quelconques}
 J. $\acute{\text{E}}$cole Polytechnique  {\bf 13}, 1--162.
\bibitem[{Oldham(1974)}]{old74} 
Oldham, K.~B. and Spanier, J. (1974).
 \emph{The Fractional Calculus},
Academic Press, New York.
\bibitem[{Ortigueira(2003)}]{ort03} 
Ortigueira, M.~D. and  Machado, J.~A.~T. (2003).
\emph{Fractional signal processing and applications},
Signal processing, {\bf{83}}, 2285-2286.
\bibitem[{Riesz(1949)}]{rie49} 
Riesz, M.  (1949). 
\emph{L'integrale de Riemann-Liouville et le probl$\acute{\text{e}}$me de Cauchy}
Acta Math.   \textbf{81},  1--223.
\bibitem[{povray(2011)}]{pov11} 
povray  (2011). 
\emph{Persistence of vision raytracer}
http://www.povray.org.
\bibitem[{Shannon(1949)}]{sha49} 
Shannon, C.~E. (1949).
\emph{Communication in the presence of noise}, 
Proc. Inst. of radio engineers,  {\bf{37}}(1), 10--21.
\bibitem[{Sparavigna(2009)}]{spa09} 
Sparavigna, A.~C. (2009).
\emph{Using fractional differentiation in astronomy}, 
arXiv.org:0910.2381.
\bibitem[{Spencer(1982)}]{spe82} 
Spencer, M. (1982).
\emph{Fundamentals of Light Microscopy}, 
Cambridge University Press.
\bibitem[{Zernike(1935)}]{zer35} 
Zernike, M.  (1935). 
\emph{Das Phasenkontrastverfaren bei der mikroskopischen Beobachtung}
Z. Tech. Phys.   \textbf{16},  454--457.
\end{thebibliography}
\end{document}